\documentclass[acmtog]{acmart}
\renewcommand\footnotetextcopyrightpermission[1]{} 
\AtBeginDocument{%
  \providecommand\BibTeX{{%
    \normalfont B\kern-0.5em{\scshape i\kern-0.25em b}\kern-0.8em\TeX}}}

\setcopyright{acmcopyright}
\copyrightyear{2019}

\usepackage{amsmath} 
\usepackage{algorithm2e}
\usepackage{graphicx}
\usepackage{subcaption}
\usepackage{caption}
\usepackage{multirow}
\usepackage{multicol}
\usepackage{slashbox}
\begin{document}

\title{Predictive modeling of brain tumor: A Deep learning approach}

\author{Priyansh Saxena}
\email{saxenapriyanshasd@gmail.com}
\orcid{0000-0003-1407-9752}

\affiliation{%
  \institution{ABV-Indian Institute of Information Technology and Management, Gwalior}
  \city{Gwalior}
  \state{Madhya Pradesh}
  \country{IN}
  \postcode{474015}
}

\author{Akshat Maheshwari}
\affiliation{%
  \institution{ABV-Indian Institute of Information Technology and Management, Gwalior}
 \city{Gwalior}
  \state{Madhya Pradesh}
  \country{IN}}
\email{aks3d76@gmail.com}

\author{Saumil Maheshwari}
\affiliation{%
  \institution{ABV-Indian Institute of Information Technology and Management, Gwalior}
\city{Gwalior}
  \state{Madhya Pradesh}
  \country{IN}}
\email{saumiliiitm@gmail.com}

\renewcommand{\shortauthors}{Priyansh Saxena, Akshat Maheshwari, Shivani Tayal and Saumil Maheshwari}

\begin{abstract}
  Image processing concepts can visualize the different anatomy structure of the human body. Recent advancements in the field of deep learning have made it possible to detect the growth of cancerous tissue just by a patient's brain Magnetic Resonance Imaging (MRI) scans. These methods require very high accuracy and meager false negative rates to be of any practical use. This paper presents a Convolutional Neural Network (CNN) based transfer learning approach to classify the brain MRI scans into two classes using three pre-trained models. The performances of these models are compared with each other. Experimental results show that the \textit{Resnet-50} model achieves the highest accuracy and least false negative rates as 95\% and zero respectively. It is followed by \textit{VGG-16} and \textit{Inception-V3} model with an accuracy of 90\% and 55\% respectively. 
\end{abstract}



\maketitle

\section{Introduction}\label{Inroduction}
With the introduction of AI techniques in the domain of e-healthcare systems, many advancements have been made in the field of medical science that aids domain experts to provide better healthcare for patients. In 2019, it was estimated that the reason for death of one out of six deaths corresponded to cancer.  Cardiovascular diseases are the principal reason for most deaths, followed by cancer. Brain tumor accounted for 85$\%$ to 90$\%$ of all primary central nervous system (CNS) tumors. According to 2019 statistics, the total annual number of deaths from brain cancer across all ages and both sexes is 247,143\cite{bib0}.

With the advancements in Deep Learning, several state of the art solutions for detecting a brain tumor exist. As a result of which tumors can be detected in the early stages and saving lives by taking preemptive measures.  A tumor can be termed as an uncontrolled cancerous tissues. The growth rate of these tissues is not normal. A tumor in the brain could be classified into one of the two types. The tumor is either \textit{benign}  or \textit{malignant}. The former has uniformity structures and contains non-active cancer cells. On the other hand, the latter has non-uniformity structures and contains active cancer cells that spread all over parts. Individuals can be affected by a brain tumor at any age, although the impact on different individuals is generally different. As per the recent data, there is a significant increase in the number of brain tumor patients. Identification of brain tumor in an earlier stage is crucial in the medical sciences for the successful treatment of a patient. Since it prevents future complications that might happen in the brain, by using the information obtained in the early stages a suitable therapy, radiation, or surgery could be suggested.

This paper presents an approach to classify brain tumor MRI images into cancerous or non-cancerous by utilizing the concepts of image processing, CNN, and transfer learning\cite{bib1}. The original image is preprocessed such that it gets converted into a standard format which is compatible with the pre-trained models used. Following this, three pre-trained models: \textit{VGG-16}\cite{bib2}, \textit{InceptionV3}\cite{bib3}, and \textit{ResNet-50}\cite{bib4} are used, and a classifier is built using a transfer learning strategy that can classify the tumor into the two classes.

The further structure of the paper is formulated into different sections. Section \ref{rel-works} illustrates the related works literature survey. Section \ref{methodology} depicts the workflow and the steps followed in the proposed approach. Section \ref{results} depicts the obtained results. Section \ref{performace-analysis} illustrates the performance analysis. Finally, Section \ref{conclusion} includes the conclusion and future scope. \\

\section{Related Works}\label{rel-works}
Magnetic resonance imaging (MRI) is a superior medical imaging technique for producing high-quality images of the parts present in the human body. It is essential to determine the correct therapy at the right stage for a tumor-infected person. The task of brain tumor segmentation from multimodal MRI scans is quite complicated. In recent time, different new methods of automated segmentation are introduced for resolving such an issue.

\textit{Beno et al.}\cite{bib5} proposed a brain tumor diagnosis system comprising of a threshold-based method used for segmentation of MR images. A collaboration of \textit{Artificial Bee Colony (ABC)} and \textit{Genetic Algorithm (GA)} was employed for predicting threshold values. \textit{El-Dahshan et al.}\cite{bib6} suggested \textit{Feedback Pulse Coupling Network (FPCNN)} for image segmentation in the brain tumor diagnosis systems. It uses \textit{Discrete Wavelet Transform (DWT)}, \textit{Principle Component Analysis (PCA)} and \textit{Artificial Neural Network (ANN)} as feature extractor and selector and classification, respectively. \textit{Emblem Ke et al.} applied \textit{SVMs} on perfusion MRI\cite{bib7} and achieved sensitivity and specificity of \textit{0.76} and \textit{0.82}, respectively. \textit{Rahmani and Akbarizadeh}\cite{bib8} proposed an unsupervised feature learning technique which was related to the collaboration of spectral clustering and sparse coding for segmentation of SAR images. \textit{Spectral clustering} is a popular image segmentation technique which makes it possible to combine features and cues. \textit{Patch-wise methods} comprise of repetitive convolutional calculations and examine spatially limited contextual features only. In order to avoid patches, a \textit{Fully Connected Network (FCN)} with deconvolution layers was used for training an end to end and pixel to pixel CNN for pixel-wise prediction with the entire image as input [10]. \textit{Chang}\cite{bib9} devised an algorithm which contained both \textit{FCN} and \textit{Conditional Random Fields (CRF)}. \textit{Ronneberger et al.}\cite{bib10} and \textit{Çiçek et al.}\cite{bib11} used \textit{UNet architecture} which consisted of two paths. These paths are namely down-sampling and up-sampling path. The former tries to capture contextual features while the latter is enables accurate localization with 3D extension. The methods in 2-dimensions ignored information about the depth. \textit{Lai}\cite{bib12} used the depth information by implementing a 3D convolution model which utilized the correlation between slices. The 3D convolution network requires a large number of parameters. Moreover, when a dataset with fewer samples is used, 3D CNN often overfit. 

In this paper, different pretrained models are applied using the transfer learning approach. These are namely \textit{VGG-16}\cite{bib2}, \textit{Inception-v3}\cite{bib3} and \textit{ResNet-50}\cite{bib4}. The main reason for choosing transfer learning is that since these models are already trained on huge datasets, so their weights can be utilized to predict the brain tumor class by simply changing the number of target classes in the output layer, which in this case is two.

\section{Methodology}\label{methodology}
The proposed methodology is organized into three different steps. This paper uses Brain MRI Images for Brain Tumor Detection dataset, which contains a total of 253 images. The dataset has 155 samples of MRI images with a malignant tumor and 98 samples benign tumor. In the first step, data-preprocessing on Brain MRI Images for tumor detection dataset\cite{bib13} is done to run and test the developed model. In the second step, data augmentation is performed since the size of the dataset is small. Finally, in the third step, different pre-trained CNN models are trained and then utilised to classify a given tumor as malignant or benign using transfer learning.\\

\subsection{Data preprocessing
}\label{data-preprocessing}
Data preprocessing can be termed as a data mining technique which involves the transformation of raw data to a format which is more interpretable and makes the images more suitable for further processing. The following preprocessing steps were involved:\\

\subsubsection{Data Splitting}\label{data-splitting}
In this step, the complete dataset is divided into 3 segments, namely, Train, Test and Validation. Train data comprises the data sample that is used for fitting the model. Validation data is that sample of the data, which helps in providing an unbiased evaluation of the model which was trained on the training data along with tuning the hyper-parameters of the model. Test data includes those samples of data which provides an unbiased evaluation of a final model that was trained on training data. The complete dataset contains 253 images, which is split into 183 training images, 50 images for the validation set, and 20 testing images.\\

\subsubsection{Crop Normalization}\label{Normalization}
It is the approach of determining of extreme points in contours. This method is used to determine the
farthest north, south, east, and west (x, y)-coordinates along a given contour. This technique applies to both raw contours and rotated bounding boxes. Using this technique, only the portion of the image containing the brain is cropped out using a Computer Vision (CV) technique given by Adrian Rosebrock\cite{bib14}. Figure 1 illustrates the aforesaid procedure.\\
\begin{figure}[hbt!]
	\begin{center}
		\includegraphics[scale=0.29]{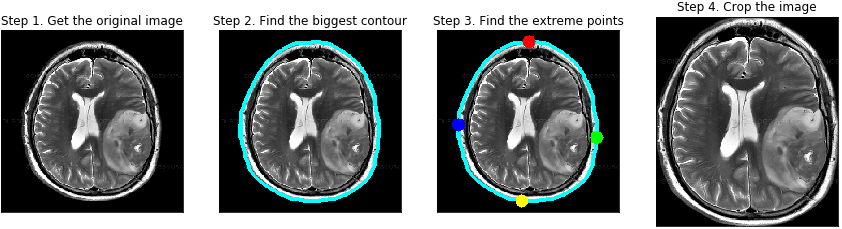}
		\caption{Finding extreme points in contours and clipping them\cite{bib14}}
	\end{center}
	\label{crop-nomal-img}
\end{figure}

\subsubsection{Resizing of images }\label{resizing}
The input dataset contains images with a different dimensions and with different aspect ratio. Therefore, the images in the dataset are resized to a preset format, since the pretrained models used require the images to be \textit{224$\times$224$\times$3} dimensions.\\

\subsection{Artificial data augmentation}\label{data-aug}
 Deep neural networks provide a significant boost in performance and produce skilful models whenever trained on more data. It is a method to artificially enhance the size of image training data by generating modified images using the original dataset. By using this technique, variations are introduced in the images which enhance the capability of the model to learn and generalize better on future unseen data. Thus by introducing variation in the training dataset, the model becomes generalized and in turn, becomes less prone to overfitting. In Figure 2, the first image s the original image present in the dataset and the remaining twenty images are generated by the technique of artificial data augmentation\cite{bib16}.

\begin{figure}[hbt!]
	\begin{center}
		\includegraphics[scale=0.31]{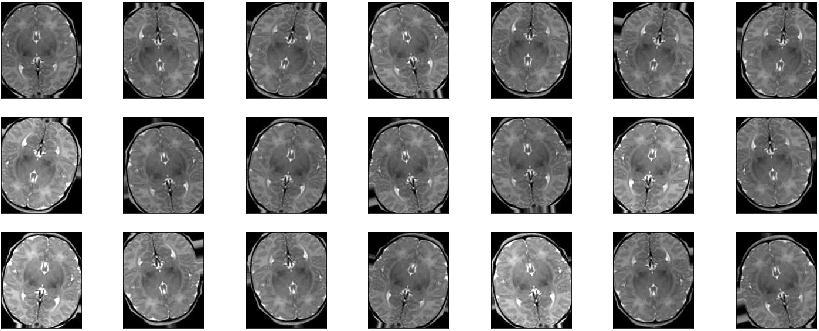}
		\caption{Data augmentation\cite{bib16}}
	\end{center}
\end{figure}

\subsection{Convolutional neural network (CNN)}
CNN's are feed-forward ANN where connections between the nodes do not form a cycle. The underlying architecture of CNN contains different layers, namely- \textit{Convolutional layer, Pooling layer} and finally, the \textit{Output layer or Fully Connected (FC) Layer}\cite{bib15}.

The convolutional layer serves to detect (multiple) patterns in multiple sub-regions in the input field using receptive fields. The pooling layer decreases the spatial size of the representation progressively and reduces number of parameters and expensive computations in ConvNet. Thus, the issue of overfitting is resolved. The inspiration behind it is that the exact location of a feature is less significant when compared to the rough location relative to other features. Max-Pooling is one of the most popular choices when designing a ConvNet. To obtain the final output, a FC layer is applied to generate an output with the required number of classes. The FC layer makes non-linear transformations on the extracted features and acts as a classifier. The generic architecture of ConvNet is presented in Figure 3.

\begin{figure}[hbt!]
	\begin{center}
		\includegraphics[scale=0.57]{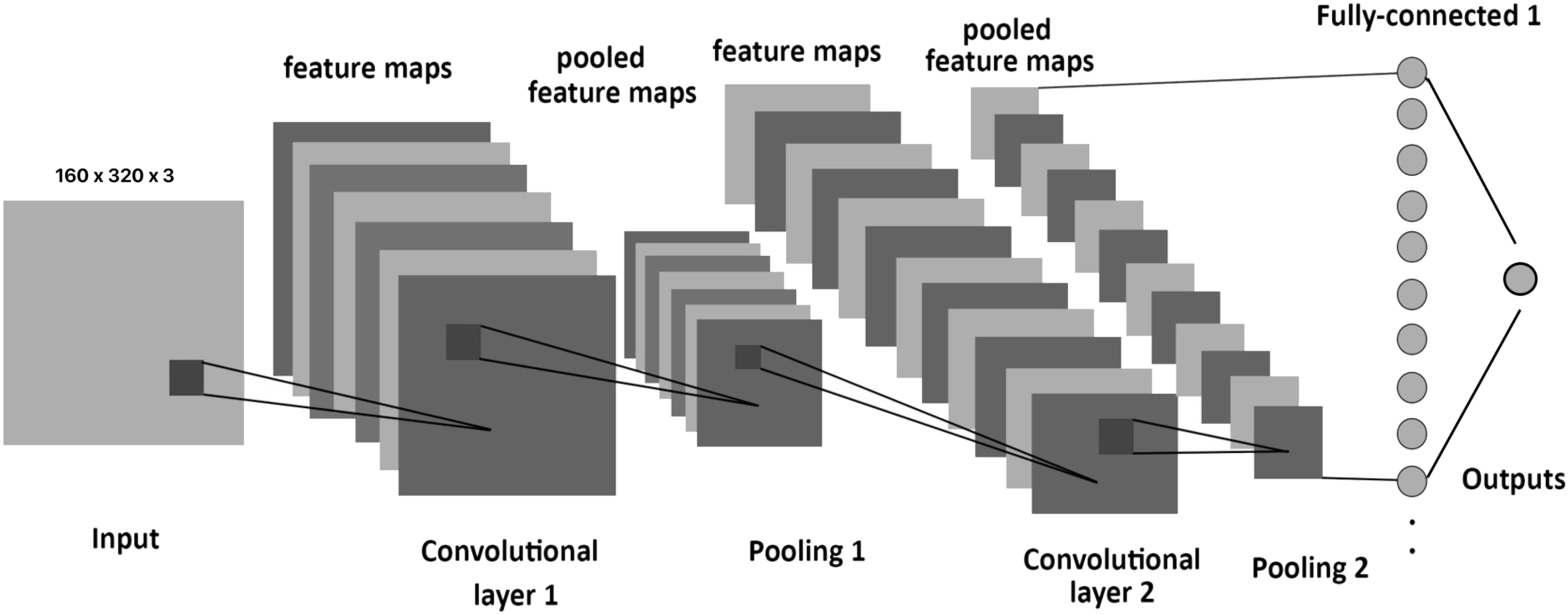}
		\caption{CNN architecture\cite{bib15}}
	\end{center}
\end{figure}

\subsection{Transfer learning}
In practice, it is improbable that an individual trains a complete CNN from scratch (with random initialization) as it is generally not possible to have a dataset with sufficient data samples. Alternately, it has become a standard procedure to pre-train a ConvNet on huge datasets like ImageNet. It contains around one million images with 1000 categories. Following that CNN can be used in two ways. The first possibility is to use it as an initialization. The second choice is to use it as a feature extractor as per the requirements. In a nutshell, Transfer learning can be considered as improved learning in a new task by transferring information learned from a existing similar problem\cite{bib1}.

\subsubsection{VGG-16}
It is a 16-layer CNN model introduced by \textit{K.
Simonyan} and \textit{A. Zisserman}\cite{bib2}. The input to first convolutional layer corresponds to a RGB image with dimensions 224$\times$224. It includes several convolutional and max-pooling layers. It also uses ReLU activation function. Model architectures prior to VGG-16 comprised of extensive kernel-sized filters. VGG-16 replaced those with multiple 3$ \times $3 filters in a serialized fashion. With a given receptive field, multiple stacked smaller size kernel is efficient when compared with a larger size kernel. The idea behind it is that the multiple non-linear layers enhances the depth of ConvNet. Thus, with an increase in depth, the ability to learn hidden features increases and that too at a lower cost. For a receptive field, multiple stacked smaller sized kernel is superior to a larger size kernel due to the extended depth of ConvNet.

\begin{figure}[hbt!]
	\begin{center}
		\includegraphics[scale=0.31]{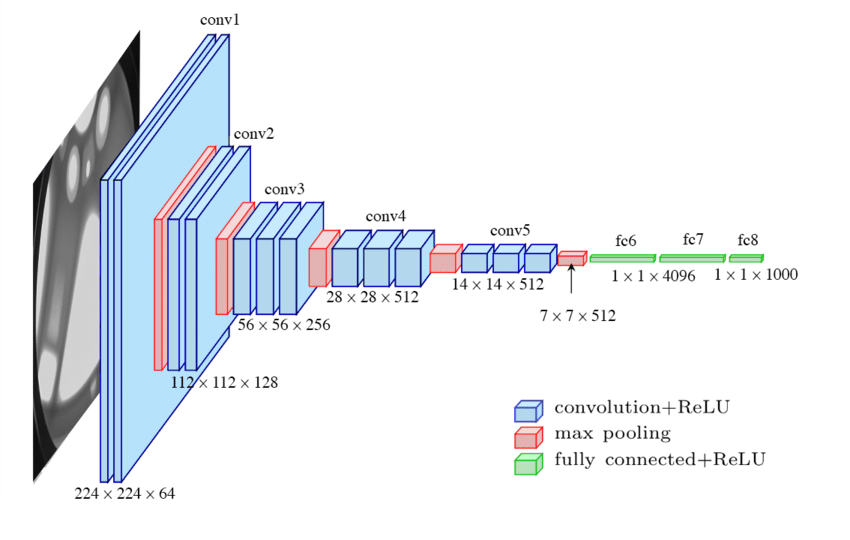}
		\caption{VGG-16 architecture\cite{bib2}}
	\end{center}
\end{figure}

\subsubsection{Inception-v3}
It is a convolutional neural network by Google Brain Team which is trained on ImageNet database. It is 48 layer deep network which classifies images into 1000 object classes. The model has capability to learn rich feature representations for different images. The network requires input image of size 299$\times$299. Inception-v3 uses batch normalization, distortion in images, RMSprop and is based on many small convolutions to significantly decrease the number of parameters.

\begin{figure}[hbt!]
	\begin{center}
		\includegraphics[scale=0.26]{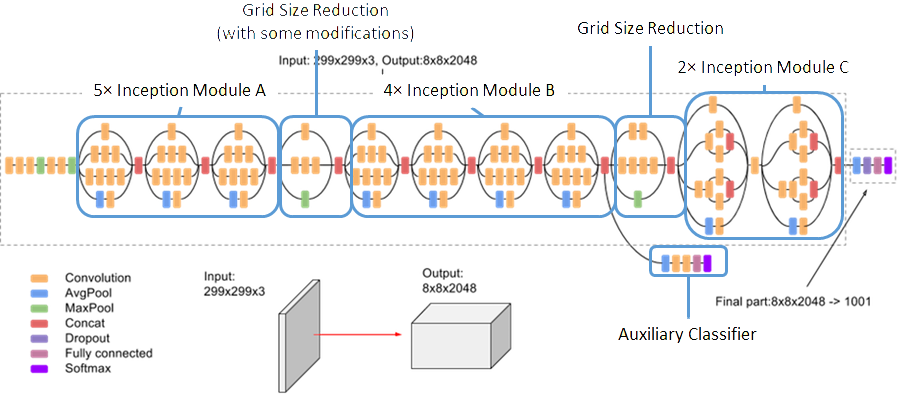}
		\caption{Inception-v3 architecture\cite{bib3}}
	\end{center}
\end{figure}

\subsubsection{Resnet-50}
It corresponds to 50 layer Residual Network by Kaiming He et al at Microsoft Research\cite{bib4}.  Residual is referred to feature subtraction. It corresponds to those features which are learned from input of that layer. ResNet performs it by using shortcut connections (directly connecting input of mth layer to some $(m+x)$\textsuperscript{th} layer). These networks are  relatively simpler to train when compared with conventional deep CNN. These also resolve the problem of degrading accuracy.

Its architecture comprises of skip connections along with massive batch normalization. These skip connections are referred to gated units or gated recurrent units. It has a lower complexity when compared with VGG.

\begin{figure}[hbt!]
	\begin{center}
		\includegraphics[scale=0.80]{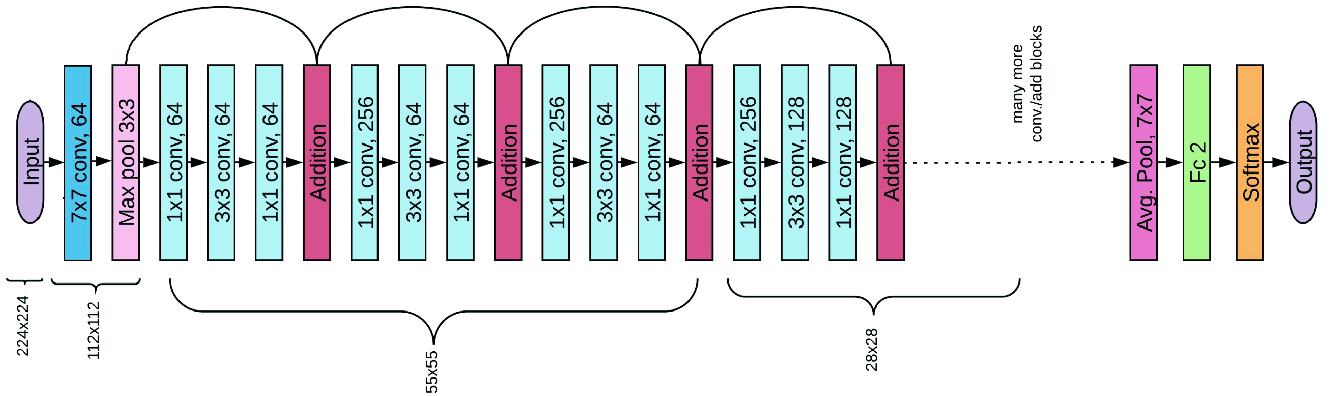}
		\caption{Resnet-50 architecture\cite{bib4}}
	\end{center}
\end{figure}

\begin{figure}[hbt!]
  \centering
  \begin{subfigure}{\linewidth}
    \centering
    \includegraphics[scale=0.28]{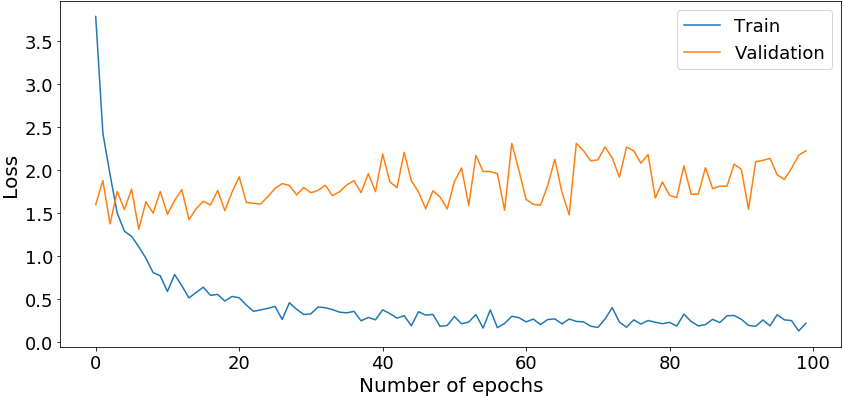}
    \caption{Training and validation loss curve}
  \end{subfigure}

  \begin{subfigure}{\linewidth}
    \centering
    \includegraphics[scale=0.28]{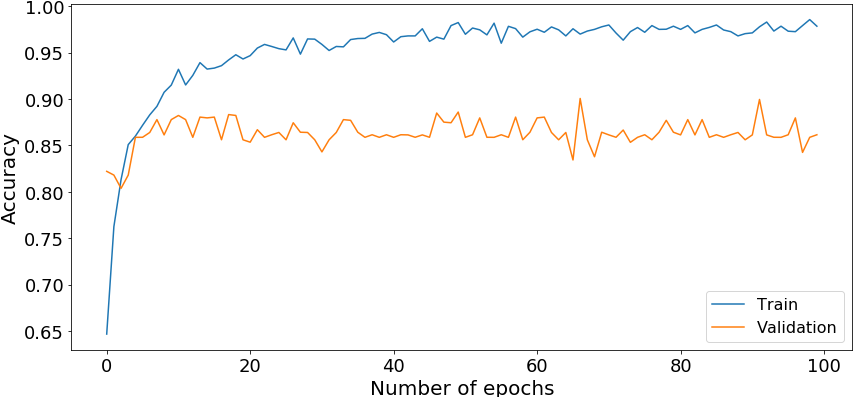}
    \caption{Training and validation accuracy curve}
  \end{subfigure}  
  \caption{VGG-16 performance metrics}  
  \label{VGG-16-results}
\end{figure}  

\section{Results and discussion}\label{results}
The experiments were performed on Brain MRI images for tumor detection dataset by Navoneel\cite{bib13}. The dataset was built by experienced radiologists using real patient's data. Data augmentation was performed with a rotation angle of 15$^{\circ}$. The experiments were performed on a Kaggle kernel, which is a cloud computational environment that enables reproducible and collaborative analysis, using 2 GPU cores, 13 GB of RAM. Each model was trained for 100 epochs. Table \ref{table1} shows the Cohen's Kappa coefficient($\kappa$), F1-score, area under the ROC curve, total training time in seconds and the accuracy of the three models used.
\noindent
\begin{table*}[htbp]
\caption{Model performance}
\begin{center}
\begin{tabular}{|l|*{5}{c|}}\hline
\backslashbox{Model}{Metric}
&\makebox[2em]{$\kappa$}&\makebox[3em]{F1-score}&\makebox[4em]{AUC-ROC}&\makebox[7em]{Training time(sec)}&\makebox[4em]{Accuracy}\\\hline
VGG-16 & 0.80 & 0.909 & 0.90  & 2158.31 & 90\%\\\hline
Inception-V3 & 0.10 & 0.689 & 0.55  & 2152.92 & 55\%\\\hline
Resnet-50 & 0.90 & 0.952 & 0.95  & 2183.72 & 95\%\\\hline
\end{tabular}\\
\label{table1}
\end{center}
\end{table*}

\begin{figure}[hbt!]
  \centering
  \begin{subfigure}{\linewidth}
    \centering
    \includegraphics[scale=0.28]{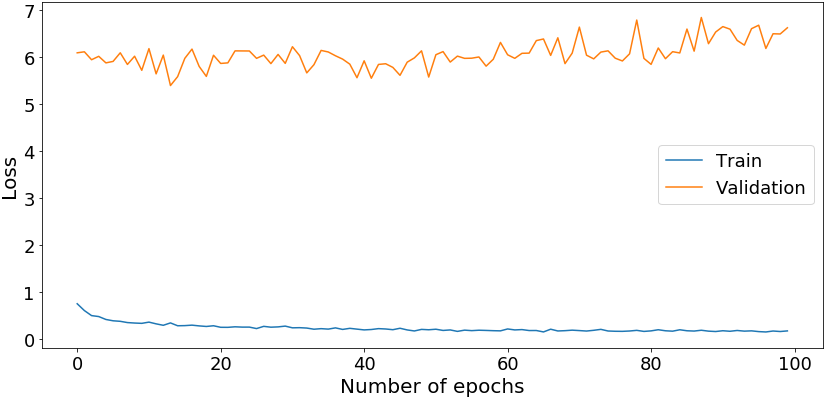}
    \caption{Training and validation loss curve}
  \end{subfigure}

  \begin{subfigure}{\linewidth}
    \centering
    \includegraphics[scale=0.28]{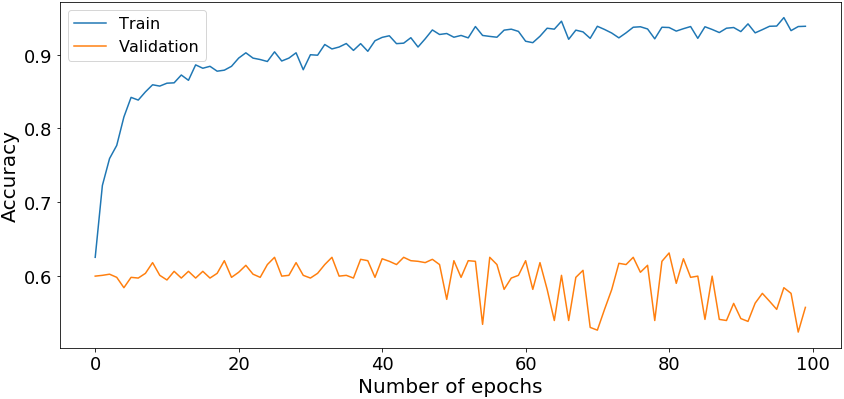}
    \caption{Training and validation accuracy curve}
  \end{subfigure}  
  \caption{Inception-V3 performance metrics}
  \label{Inception-V3-results}
\end{figure}

\begin{figure}[hbt!]
  \centering
  \begin{subfigure}{\linewidth}
    \centering
    \includegraphics[scale=0.28]{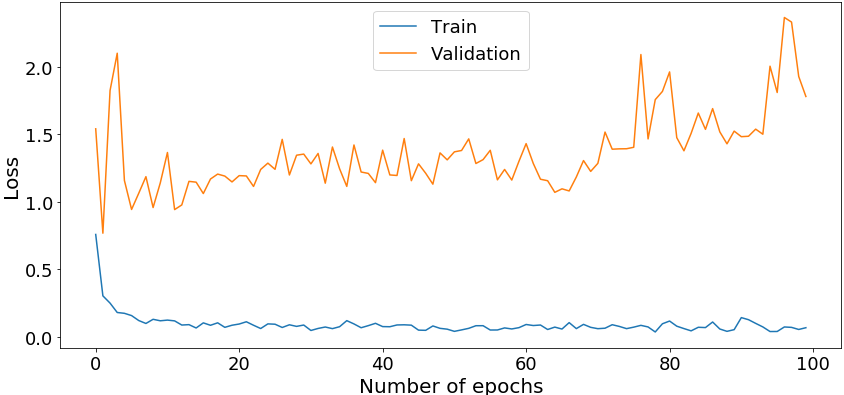}
    \caption{Training and validation loss curve}
  \end{subfigure}

  \begin{subfigure}{\linewidth}
    \centering
    \includegraphics[scale=0.28]{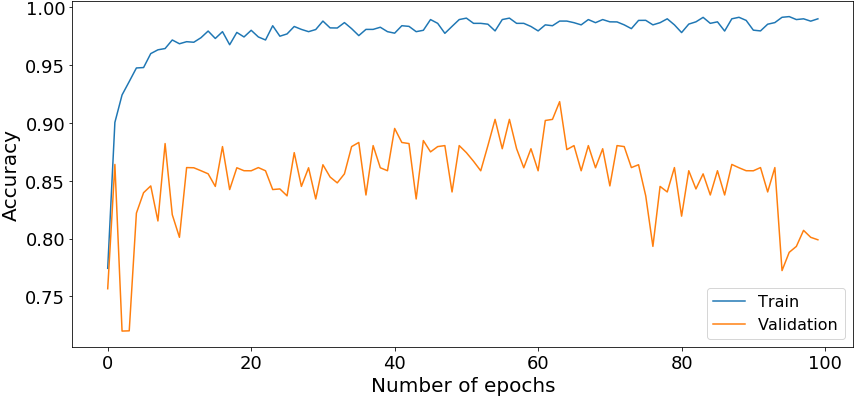}
    \caption{Training and validation accuracy curve}
  \end{subfigure}  
  \caption{Resnet-50 performance metrics}
  \label{Resnet-50-results}
\end{figure}

Figures \ref{VGG-16-results} to \ref{Resnet-50-results} were obtained by training the different models on the Brain MRI images dataset for 100 epochs and using checkpointing. Plots (a) and (b) of Figures \ref{VGG-16-results} to \ref{Resnet-50-results} shows the loss and accuracy for training and validation phase, respectively for the three models.

\section{Performance Analysis}\label{performace-analysis}
 The performance for the proposed methodology was measured in terms of Cohen's kappa($\kappa$), F1-score, area under the ROC curve (AUC-ROC) and test accuracy. Resnet-50 had the highest F1-score due to high values of precision and recall (F1-score is harmonic mean of precision and recall). Along with accuracy, the false-negative rate is considered of the highest importance as misclassifying a cancerous patient as safe could lead to loss of life of an individual. Resnet-50 had zero false-negative rate on the test data, which makes it suitable for practical purposes. The area under the ROC curve is highest for Resnet-50 and least for inception-V3. The performance of Inception-v3 was very close to a random classifier since its AUC is 0.55. Resnet-50 achieves the highest test accuracy while the inception-V3 suffered from overfitting. Cohen's kappa score is again least for inception-v3 and highest for Resnet-50. Thus, by analyzing and comparing the overall effect of all performance metrics, it can be seen that Resnet-50 has outperformed the other two models in brain tumor classification.

\section{Conclusion and future scope}\label{conclusion}
In this paper, three pre-trained CNN models using transfer learning technique are applied to classify brain tumor MRI images into malignant and benign. After preprocessing the dataset, the MRI images are cropped out of dark corners using crop normalization technique. Following that data augmentation is performed to introduce variations in the data samples by increasing the size of training sets which in turn helped to create a generalizable model and thus reducing overfitting. Resnet-50 had the best F1 score and highest test accuracy with the zero false-negative rate, which is the primary concern in an automatic diagnostic system and thus making it suitable for practical applications. The area under the ROC curve and Cohen's Kappa coefficient($\kappa$) are also highest for Resnet-50 model. Inception-V3 suffered from overfitting and is just slightly better than a random classifier with an accuracy 0.55. By using the transfer learning and data augmentation technique, it was possible to achieve such high accuracy even on such smaller datasets. The future possibilities could be to use a larger dataset to increase the accuracy further or to use a new state of the art architectures by using transfer learning technique. Extensive hyper-parameter tuning and a better preprocessing technique can be devised to enhance the model performance further.



\end{document}